\documentclass[11pt,a4paper]{article}
\usepackage{caption}
\usepackage{graphicx}
\usepackage{tikz}
\usepackage{bm}

\def\checkmark{\tikz\fill[scale=0.4](0,.35) -- (.25,0) -- (1,.7) -- (.25,.15) -- cycle;} 
\usepackage{dsfont}
\usepackage{algorithm}
\usepackage{multicol}
\usepackage{algpseudocode}
\usepackage{booktabs}
\usepackage{enumitem}

\usepackage{mathtools}
\DeclarePairedDelimiter\ceil{\lceil}{\rceil}

\usepackage{hyperref}
\usepackage{url}
\usepackage{times}
\usepackage{latexsym}
\usepackage{times,latexsym}
\usepackage{url}
\usepackage[T1]{fontenc}

\usepackage[acceptedWithA]{tacl2021v1}

\usepackage{xspace,mfirstuc,tabulary}

\newif\iftaclinstructions
\taclinstructionsfalse 
\iftaclinstructions
\renewcommand{\confidential}{}
\renewcommand{\anonsubtext}{(No author info supplied here, for consistency with
TACL-submission anonymization requirements)}
\newcommand{\instr}
\fi

\iftaclpubformat 

\else

\fi

\title{DP-Parse: Finding Word Boundaries from Raw Speech \\with an Instance Lexicon}

\author{Robin Algayres$^{1,2,4}$ Tristan Ricoul$^{2}$ Julien Karadayi$^{2}$ Hugo Laurençon$^{2}$ Salah Zaiem$^{2}$ \\
\textbf{Abdelrahman Mohamed}$^{1}$ \textbf{Benoît Sagot}$^{3}$ \textbf{Emmanuel Dupoux}$^{1,2,3,4}$\\
Meta AI Research$^1$, ENS/PSL,Paris$^2$, 
EHESS, Paris$^3$, Inria, Paris$^4$\\
\texttt{\{robin.algayres, benoit.sagot\}@inria.fr} \\
\texttt{dpx@fb.com}
}
\date{}

\begin{document}

\maketitle

\begin{abstract}
Finding word boundaries in continuous speech is challenging as there is little or no equivalent of a `space' delimiter between words. 
Popular Bayesian non-parametric models for text segmentation \cite{goldwater2006contextual,goldwater2009hdp} use a Dirichlet process to jointly segment sentences and build a lexicon of word types. We introduce DP-Parse, which uses similar principles but only relies on an \textit{instance lexicon} of word tokens, avoiding the clustering errors that arise with a lexicon of word types.
On the Zero Resource Speech Benchmark 2017, our model sets a new speech segmentation state-of-the-art in 5 languages. 
The algorithm monotonically improves with better input representations, achieving yet higher scores when fed with weakly supervised inputs. Despite lacking a type lexicon, DP-Parse can be pipelined to a language model and learn semantic and syntactic representations as assessed by a new spoken word embedding benchmark. \footnote{Our open-source implementation of DP-Parse: \href{https://gitlab.cognitive-ml.fr/ralgayres/dp-parse}{https://gitlab.cognitive-ml.fr/ralgayres/dp-parse}}
\end{abstract}
\vspace{-1.em}
\section{Introduction}
One of the first tasks that infants face is to learn the words of their native language(s). To do so, they must solve a word segmentation problem, since words are rarely uttered in isolation but come up in multi-word utterances \cite{brent}. Segmenting utterances into word or subword units is also an important step in NLP applications, where an input string of orthographic symbols is \textit{tokenized} into words or sentence piece before being fed to a language model. While many writing systems use white spaces that make (basic) tokenization a relatively simple task, others do not (e.g. Chinese) and turn tokenization into a challenging machine learning problem \cite{liyuan1998chinese}. A similar situation arises for `textless' language models based on units derived from raw audio \cite{lakhotia2021gslm}, which, like infants, do not have access to isolated words or word-separating symbols.  


The most successful approach to unsupervised word segmentation for text inputs is based on \textit{non-parametric Bayesian models} \cite{goldwater2006contextual,goldwater2009hdp,snlm,kamper2017eskmeans,berg2010,ag_ext,johnson2007ag,ag_the} that jointly segment utterances and build a lexicon of frequent word forms using a Dirichlet process. The intuition is that frequent word forms function as anchors in a sentence, enabling the segmentation of novel words (like `dax' in `did you see the dax in the street?'). Such models are tested on \textit{phonemized texts} obtained by converting text into a stream of phonemes after removing spaces and punctuation marks. Even though phonemized text may seem a reasonable approximation of continuous speech, such models have given disappointing results when applied directly to speech inputs. Since the direct comparison between speech-based and text-based models was introduced in 2014 by \citet{ludusan2014metrics}, the gap of performance between these two types of inputs as documented in three iterations of the Zero Resource Speech Challenge focused on word segmentation has remained large \cite{Versteegh2016,dunbar2017,dundar2020zs20}. 
We attribute these difficulties to two major challenges posed by speech compared to text inputs: acoustic variability and temporal granularity, which we address in our contribution. 

The first and most important challenge is \textit{acoustic variability}. In text, all tokens of a word are represented the same. In speech, each token of a word is different, depending on background noise, intonation, speaker voice, speech rate, etc. Text-inspired speech segmentation algorithms \cite{kamper2017eskmeans} apply a clustering step to word tokens in order to get back to  word types. We believe that this step is unnecessary and is responsible for the low performance of existing systems, as errors in the clustering step are not recoverable and negatively impact  word segmentation. We propose an algorithm that segments speech utterances based on an \textit{instance lexicon} of word tokens, instead of a lexicon of word types. Each speech segment instance of the training set is represented as a distinct memory trace, and these traces are used to estimate word form frequency using a $k$-NN algorithm without having to refer to a discrete word type but still applying the same Dirichlet process logic. As for the representation of these word tokens, we follow recent approaches that use fixed length \textit{Speech Sequence Embeddings (SSE)} 
\cite{thual2018knn,kamper2017eskmeans}. We use state-of-the-art SSEs from \citet{algayres2022sse} that have either been trained in a self-supervised fashion or with weak labels, in order to assess how the segmentation model behaves with inputs of increasing quality.

The second challenge is related to the fact that the speech waveform being \textit{continuous} in time, the number of possible segmentation points grows very large, making the optimization of segmentation of large speech corpora using Bayesian techniques \textit{intractable}. Some models reduce the number of segmentation points using phoneme boundaries \cite{lee2012nonparametric,bathi2021scpc}, syllable boundaries \cite{kamper2017eskmeans} or a constant discretization of the time axis into speech frames, subject to the following trade-off: too coarse frames may miss some short word boundaries, too fine-grained ones may render segmentation intractable on large corpora because of the quadratic increase in number of segmentations with frame rate. Here we choose a compromise with 40ms frames, corresponding to half of the mean duration of a phoneme, yielding a 4x theoretical slowdown compared to phoneme-based transcripts. In addition, we introduce an \textit{accelerated version} of a Dirichlet process segmentation algorithm that replaces Gibbs sampling of each boundary by sampling entire \textit{parse trees} using Dynamic Programming Beam Search, and updating the lexicon in \textit{batches} instead of continuously. This achieves a 10 times speed-up compared to \citet{johnson2007ag}'s implementation, with similar or superior performances. 

Because of our `no clustering' approach, we cannot evaluate our model using the word type-based metrics of the Zero Resource speech segmentation challenge \cite{Versteegh2016, dunbar2017zerospeech}. Here, we use the two classes of metrics: boundary-related metrics  (Token and Boundary F-score) that do not refer to types, and high-level word embedding metrics.  The idea is use a downstream language model to learn high level word embeddings and evaluate them on semantic and Part-Of-Speech (POS) dimensions. We rely on a semantic similarity task from \cite{dundar2020zs20} and introduce two new metrics. 

The combination of our contributions yields an overall system, DP-Parse, that sets a new state-of-the-art on speech segmentation for five speech corpora by a large margin. By doing various ablation and anti-ablation studies (replacing unsupervised components by weakly supervised ones) we pinpoint the components of the system where there is still margin for improvement. In particular, we show that using better embeddings obtained with weak supervision, it is possible to double the segmentation F-score and substantially improve our new semantic and POS scores. 

\section{Related Work}\label{sec:related}

\subsection{Speech Sequence Embeddings}

Speech Sequence Embedding (SSE) models take as input a piece of speech signal of any length and outputs a fixed-size vector. The main objective of these models is to represent the phonetic content of a given speech segment. A naive SSE model would be to extract frame-level features of a speech sequence, using for instance Wav2vec2.0 or HuBERT \cite{baevski2020w2v2,ning2021hubert}, and mean-pool the frames along the time axis. A more subtle approach is to train a self-supervised systems on positive pairs of speech sequences. The model learns speech sequence representation thanks to contrastive loss functions (e.g. the infoNCE loss as in \citet{algayres2022sse,jacobs2021multilingualsse} or the Siamese loss as in \citet{settle2016siamese,riad2018siamese}) or reconstructive losses (e.g. \textit{Correspondance Auto-Encoder} from \citet{kamper2018cae} where the first element of the pair is compressed and decoded into the second element). Recently, a growing body of work leverage multilingual labelled dataset to build SSEs that can, to an extent, generalize to unseen languages \cite{hu2020multilingualsse,hu22020multilingualsse,jacobs2021multilingualsse}. In this work, we use a state-of-the-art self-supervised system SSE model from \citet{algayres2022sse}.

\subsection{Speech segmentation}

Three main class of models have been proposed to solve speech segmentation: \textit{matching-first} models \cite{park2007unsupervised,jansen2011dtw,garcia} attempt to find high quality pairs of identical segments and cluster them based on similarity, thereby building a lexicon of word types which may not cover the entire corpus \cite{bhati2020selfexpressing,okko2015sylseg}. \textit{Segmentation-first} models try to exhaustively parse a sentence \cite{lee2015lexicon,kamper2016segmentation,kamper2017eskmeans,kamper2020vaewordseg}, while jointly learning a lexicon (and therefore also involving matching and clustering). \textit{Segmentation-only} models discover directly the likely word boundaries in running speech using prediction error \cite{bathi2021scpc,cuervo2021acpc}, without relying on any lexicon of word types. Our approach is a new hybrid of the last two lines of research: like segmentation-first model, we jointly model lexicon and segmentation, but our lexicon is not a lexicon of types, thereby escaping  matching and clustering errors, like in segmentation-only models. This model shows good performances across different languages as measured by the Mean Average Precision on pre-segmented spoken words.

\subsection{Evaluating speech segmentation}

Across the different word segmentation studies, the evaluation has been done according to two general classes of metrics: matching and clustering metrics for word embeddings (MAP, NED, grouping F-score, Type F-score), and segmentation metrics for boundaries (Token and Boundary F-scores) \cite{carlin2011rapid,ludusan2014metrics}. All of these metrics presuppose that the optimal segmentation strategy is aligned with the text-based gold standard (based on spaces and punctuation). Yet, it is entirely possible that such segmentation is not optimal for downstream applications, as witnessed by the fact that most current language models use subword units like BPEs or sentence pieces \cite{devlin2018bert}. Therefore, we propose a third class of segmentation metrics  based on downstream language modelling tasks. \citet{glass2018speech2vec} showed that gold word boundaries provide sufficient information to learn good quality semantic embeddings from raw audio using a skipgram or word2vec objective, as assessed by semantic similarity benchmarks adapted to speech inputs. These datasets were also used to compile the sSIMI benchmark in the Zero Speech Challenge 2021 \cite{tuanh2021zerospeech}, but turns out to give very low and noisy results (at least for small training sets). Here, we introduce two new metrics evaluating semantic and grammatical similarity that we show to be more stable.

\begin{figure}[]
\includegraphics[width=1\linewidth]{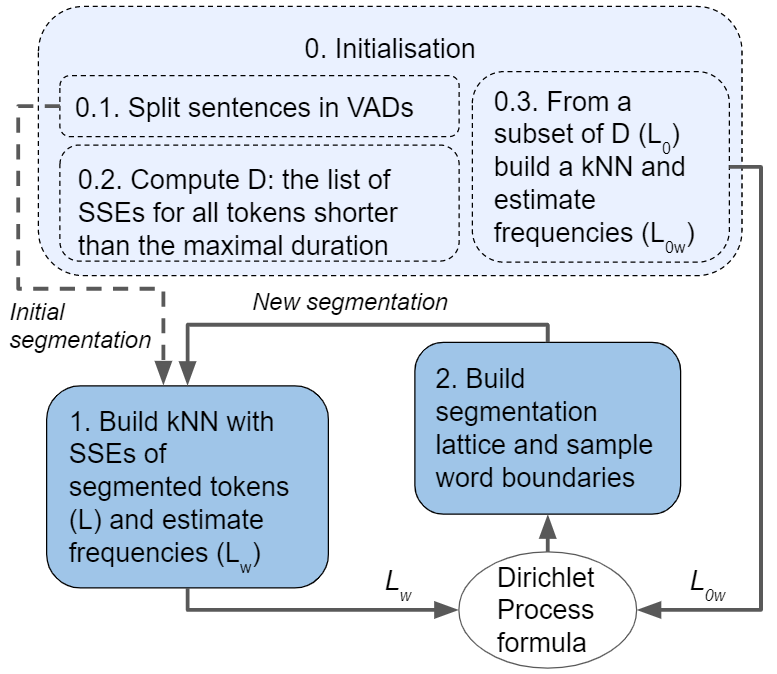}
\caption{\textbf{The core components of DP-Parse}. The algorithm is a loop between two main steps (Step 1. and 2.) in the spirit of the Expectation Maximization algorithm. Given an initial coarse word segmentation, we estimate $L_w$ and $L_{0_w}$ (i.e. the Dirichlet Process parameters) with k-NN density estimation over SSEs. The estimated parameters are used to derive a new word segmentation, which in turn serves to re-estimate $L_w$ and $L_{0_w}$. }
\label{fig:dpparse}

\end{figure}

\section{Method: Instance-based Dirichlet Process Parsing}\label{sec:dirichletparsing}

This section describes the building blocks and notations used by DP-Parse as depicted on Figure~\ref{fig:dpparse} and Table~\ref{notation_table}, respectively. At a high level, DP-Parse is a new adaptation of Goldwater's Dirichlet process algorithm \cite{goldwater2009hdp} for speech inputs.
At the heart of Goldwater's algorithm, is a hierarchical Bayesian model which generates a joint distribution of a lexicon of words (a word is a sequence of phonemes) and a corpus (a sequence of words). The associated learning algorithm optimizes the likelihood of an observed corpus as a function of the (unobserved) lexicon and corpus segmentation. It does so by alternating between two EM steps: (Step 1) estimating the most probable lexicon given a segmentation, and (Step 2) sampling among the most probable segmentation given the lexicon. More precisely, at each Step 1, the algorithm estimates the probability of a segmentation from two sets of numbers: the probability $P^0_W{(w)}$ that a given speech segment $w$ is a word, and $L_w$ the frequency of a given word $w$ in the (segmented) input corpus. They are estimated by building tables of counts of phoneme-ngrams and words given a segmentation, respectively. At Step 2, these numbers are combined using Equation \ref{eq:dp} (Section \ref{sub:formula}) and a new segmentation is sampled. 
Our adjustments to the original algorithm are the following: 

\begin{itemize}[leftmargin=*]
  \setlength\itemsep{-0.1em}
    \item Section \ref{sub:embeddings}: Phonemes are replaced by 40ms speech frames (from a pretrained self-supervised learning algorithm), words are replaced by fixed-length \textit{speech sequence embeddings} (using a pretrained self-supervised embedder).
    \item 
    Section \ref{sub:dens}: The tables of counts for computing  $P^0_W{(w)}$ and $L_w$ are replaced by k-Nearest-Neighbors ($k$-NN) indexes of embeddings ($L_0$, and $L$, respectively) which can be viewed as instance lexicons of all possible n-frames and of segmented `words', respectively. The count estimates are obtained using Gaussian kernel \textit{density estimation} over $k$-NN. 
    \item Section \ref{sub:formula}: We make a small adaptation to the\textit{ Dirichlet process formula}.
    \item Section \ref{sub:lattice}: Instead of alternating between the two steps at each boundary as in the original Gibbs sampling algorithm, which is very time-consuming, we sample segmentations over entire utterances using a \textit{segment lattice}, and update the lexicon over a batch of utterances. 
    \item Section \ref{sub:init}: We \textit{initialize} the system by selecting as an initial lexicon a list of short enough utterances, and precompute all the constant values ($L_0$ and $P^0_W{(w)}$) to optimize for speed. 
\end{itemize}

So modified, the algorithm is no longer dependent on input type and can work either with discrete representations (text, represented as sequences of 1-hot vectors of phonemes) or continuous ones (speech, represented as embeddings).

\subsection{Speech Sequence Embeddings}\label{sub:embeddings}

We represent speech as 20ms frames obtained by selecting the 8th layer of a pretrained Wav2vec2.0 Base system from \citet{baevski2020w2v2}. Each two successive frames are tied together so that a speech sentence is a series of 40ms speech blocks. To represent a speech segment, we use the Speech Sequence Embedding (SSE) model from \citet{algayres2022sse}: a self-supervised system trained with contrastive learning where positive pairs are obtained by data-augmentation of the speech signal. This model shows good performances across different languages as measured by the Mean Average Precision on pre-segmented spoken words. The SSE model takes as input the pre-extracted Wav2vec2.0 frames and apply a single 1-D convolution layer with GLU (kernel size: 4, number of channels: 512, stride: 1) and a single transformer layer (attention heads: 4, size of attention matrices: 512 neurons, and feed forward layer: 2048 neurons ). A final max-pooling layer along the time axis is applied to get a fixed-size vector. To save computation, we reduce the dimensionality of \citet{algayres2022sse}'s SSE model from 512 to 64 with a PCA trained on random speech segments extracted from the corpus at work.

In addition to their unsupervised SSE model, \citet{algayres2022sse} provides a weakly-supervised version. They trained it with the same contrastive loss (infoNCE) using positive pairs obtained with a time-aligned transcription of the speech signal. The weakly-supervised version will be used as a topline model, as it scores much higher on Mean Average Precision than the self-supervised one.

\begin{table*}[]
\centering
\resizebox{\textwidth}{!}{%
\begin{tabular}{ll}
\hline
$u$                                 & a unit $u$ is a block of 40 ms of speech signal \\ \hline
$w=(u_0,...,u_w)$                   & a segment $w$ is a sequence of units                    \\ \hline
$E_{w} \in {\rm I\!R}^p$            & a fixed length embedding of a segment $w$ computed by a SSE model                    \\ \hline
$C=\{v^0,...,v^N\}$                   & a corpus $C$ is a set of utterances (segments without silence found by VAD) \\ \hline
$C_{seg}=\{(w^0_1,...,w^0_{p_0}), ...\}$ & a segmented corpus $C_{seg}$ is a set of sequences of segments    \\ \hline
$D=\{w_0,...w_{P}\}$              & the list of all possible subsegments in  $C$ that are possible words (between 40ms and 800ms) \\ \hline
$L_0=\{E_{w_0},...,E_{w_{P}}\}$              & a kNN index of  embeddings from $D$   \\ \hline
$L=\{E_{w_0},...,E_{w_n}\}$                    & a kNN index of all embeddings from the segments in $C_{seg}$                                             \\ \hline
$L_{0_w}$                           & the frequency of $E_w$ in $L_0$                            \\ \hline
$L_w$   & the frequency of $E_w$ in $L$ \\ \hline
\end{tabular}
}
\caption{Notations used in Section \ref{sec:dirichletparsing}}
\label{notation_table}
\end{table*}

\subsection{Density estimation from Gaussian-smoothed k-Nearest-Neighbors}\label{sub:dens}
\label{densityestimation}

In text, the exact frequency of a string of letters can be computed by counting how often it occurs in the corpus. In speech, due to acoustic variability, most of the counts would be 1, and clustering methods cause too many errors to get reliable count estimates (more details in appendix \ref{appendix:kmeans}). Here, we follow the method in \citet{algayres2020sse} using Gaussian filtering over $k$-NNs. 
Let us assume that all speech segments $w$ of a corpus are represented as embeddings $E_w$ and stored in a $k$-NN index $L_0$. To compute $L_{0_w}$ the frequency of $w$ in $L_0$, we extract the $k$ nearest-neighbors of $E_w$: $(E_1,...,E_k)$\footnote{We remove from the list the embeddings corresponding to overlapping segments} and sum over them, weighted by a decreasing function of their distance to $w$. Intuitively, embeddings close to $E_w$ are more likely to be instances of the same segments than distant ones.
For weighting, we use a Gaussian kernel centred on $E_w$ as in the following equation: 
\vspace{-0.7em}
\begin{equation}
L_{0_w} \approx F(w)= \sum_{j=1}^{k} \exp^{-\beta \lVert E_{w}-E_{j} \lVert^2_2}
\label{eq:density}
\end{equation}

$F$ is the Parzen-Rosenblatt window method for density estimation \cite{parzen} and returns a continuous value between $0$ and $k$, estimating $L_{0_w}$. It rests on two hyper-parameters: $k$ which is set to $1000$ and $\beta$, the standard deviation of the Gaussian kernel. To set $\beta$, we follow the observation from \citet{algayres2022sse}: around 50\% of the segments in the development dataset of the Zerospeech Challenge 2017 appear to have a frequency of one. Therefore, we set $\beta$ during runtime so that 50\% of the $L_{0_w}$ calculated get a frequency of one (i.e. an F value below a small $\epsilon>0$). We did not change these hyper-parameters for any of the tested languages.

\subsection{Dirichlet Process formula}\label{sub:formula}

This section explains how \citet{goldwater2009hdp} formulates the Dirichlet Process and our modification.
Let $v'$ be a non silent section from a speech corpus $C$ found by a Voice Activity Detection (VAD). A segmentation of $v'$ is written $v'_{seg}$ and is composed of a series of segment $(w_1,...,w_l)$. The probability of $v'_{seg}$ is, under the unigram hypothesis, the product of the probability of each segment to be a real spoken word: 
\begin{equation}
P_S(s'_{seg})=\prod_{i=0}^{l}P_W(w_i) 
\label{pseg}
\end{equation}
To model the probability of a segment $w$ to be a real word, \citet{goldwater2009hdp} uses the following formulation of the Dirichlet Process:
\begin{equation}
    P_W(w|C,C_{seg})=\frac{L_w}{\#L+\alpha_0}+\frac{\alpha_0P^0_W(w)}{\#L+\alpha_0}
    \label{eq:dp}
\end{equation}
The first term of Equation \ref{eq:dp} accounts for how often the token $w$ has been segmented in $C_{seg}$. Its numerator $L_w$ is the count estimate of $w$ in $L$. The second part of Equation \ref{eq:dp} is the intrinsic probability of $w$ to be a word regardless of its appearance in $C_{seg}$. This intrinsic probability is called the base distribution $P^0_W$ and is controlled by the concentration parameter $\alpha_0$.\\

To find a formulation for $P^0_W$, the intuition from \citet{goldwater2009hdp} is simple: frequently appearing tokens are more likely to be true words than rare ones. Yet, their formulation of $P^0_W$ cannot be easily adapted to the segmentation of speech data. Our contribution here is to propose a new formula for $P^0_W$ which follows the same intuition: 
\begin{equation}
P^0_W(w)=\frac{L_{0_w}}{\#L_0}
\label{eq:dpp0}
\end{equation}
where $L_{0_w}$ is the count estimate of $w$ in $L_0$ (the lexicon of all possible segments in $C$). $\#L_0$ being the total number of segments in $L_0$, $P^0_W$ is then the probability to find $w$ in among the tokens in $C$. 

\subsection{Segmentation lattice}\label{sub:lattice}

In \citet{goldwater2009hdp}, the learning algorithm uses Gibbs sampling, where word boundaries are sampled one at a time, requiring all the parameters of the Dirichlet Process model to be recomputed after each sample. On text data, it is not a problem, as updating the model's parameters ($L_w$ and $L_{0_w}$) is fast: no $k$-NN search is needed, parameters are computed exactly by matching and counting strings. It becomes a bottleneck for speech segmentation where the parameters are computed with density estimation and $k$-NN search. One way to alleviate this problem, is to incorporate the dynamic programming version of the Gibbs sampler from \cite{mochihashi2009bayesian}. Another way, as in this paper, is to build a segmentation lattice over each utterance and sample among the N-best segmentations.\\ 

Here, we assume that the corpus $C$ has been pre-segmented in a set of utterances using a VAD algorithm. After each Step 1 from Figure \ref{fig:dpparse} we can compute the log probability, $\log(P_W(w))$ from Equation \ref{eq:dp}, that each token in an utterance is a real word. Yet, instead of directly using this probability, we introduce a per-token penalty score that favors short tokens:\begin{equation}
q(x)=\left( \frac{x-1}{\delta}\right)^{\gamma}
\label{eq:q}
\end{equation}
This penalty is added to the dirichlet process log probability as follows:
\begin{equation}
S(w)= \log(P_W(w)+\epsilon)+q(len(w))
\label{eq:penalty}
\end{equation}
where $\epsilon$ is a very small number to handle cases where $P_W(w)=0$, $len(w)$ is the number of 40ms speech frames in $w$.

\begin{figure}
    \centering
    \includegraphics[width=\linewidth]{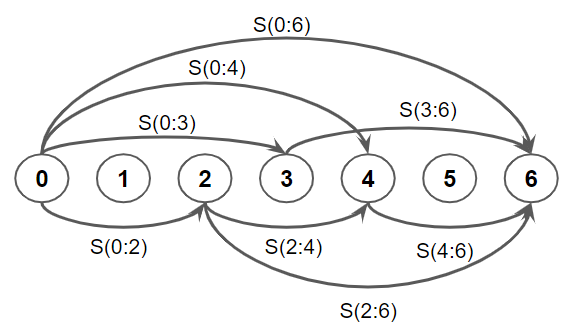}
    \caption{An example of a segmentation lattice for a small utterance of 6 units, with a constraint on word tokens to be bounded between 2 and 6 units. A segmentation path is a sequence of segmentation arcs that covers the whole utterance. Each arc starts and ends in-between units and is associated with its score $S$ from Equation \ref{eq:penalty}}
    \label{table:lattice}
\end{figure}

For each utterance in $C$, we create a segmentation lattice that provides a compact view for all possible segmentation paths. A segmentation path is a sequence of consecutive segmentation arcs that covers the full utterance. Each arc starts and finishes in-between the units and is bounded within a minimal and maximal length. 
An example of a segmentation lattice can be found at Table \ref{table:lattice}. Each segmentation arc is associated with its $S$ score from Equation \ref{eq:penalty}. For each utterance, the N-best segmentations are computed with Dynamic Programming Beam Search, and we sample from a softmax of their total $S$ scores. 
One advantage of this procedure is that it is possible to parallelize utterance segmentation across large batches and only update the lexicon $L$ after each batch. In our experiments, we take the entire corpus $C$ to be a single batch.

\subsection{Initialising DP-Parse}\label{sub:init}

We create a corpus $C$ as a collection of utterances by applying the pyannote VAD \cite{bredin2019pyannote} with a threshold of 200 ms. 
As shown in Figure \ref{fig:dpparse}, initialization (Step 0) contains several sub-steps. The first (0.1) is to provide an initial segmentation of $C$, using a simple heuristic: all sentences shorter than 800ms are treated as word tokens, the other ones are discarded. The intuition is that short sentences could be words in isolation, which provides the seed for an initial lexicon.

The second sub-step (0.2) is to create the list $D$ composed of the embeddings of all possible speech segments in $C$ that are possible words, which we set to be anything between 40ms and 800ms (by 40ms increments). To embed a speech segment, we use the SSE model in Section \ref{sub:embeddings}. 

The third sub-step (0.3) consists in the construction of $L_0$: a $k$-NN index of all embedded segments in $D$. In practice, we found that randomly subsampling $D$ to one million embedded segments worked well (see grid-search in appendix \ref{appendix:hp}), and we precompute $L_{0_w}$ as explained in Section \ref{sub:dens}.

\section{Evaluation metrics and settings}

\subsection{Boundary level segmentation metrics}

In the Zerospeech Challenge 2017 \cite{dunbar2017zerospeech}, two metrics measure how well discovered boundaries matches with the gold word boundaries. These metrics, the Boundary and Token F-score, are obtained by comparing the discovered sets of tokens to the gold one obtained by force-aligning spoken sentences with their transcription. The evaluation is done in phoneme space and each discovered boundary is mapped to a real phoneme boundary: if a boundary overlaps a phoneme by more than 30ms or more than half of the phoneme length, the boundary is placed after that phoneme (otherwise it is placed before).

\subsection{Semantic and POS embedding metrics}

\begin{figure}
     \begin{center}
     \includegraphics[width=\linewidth]{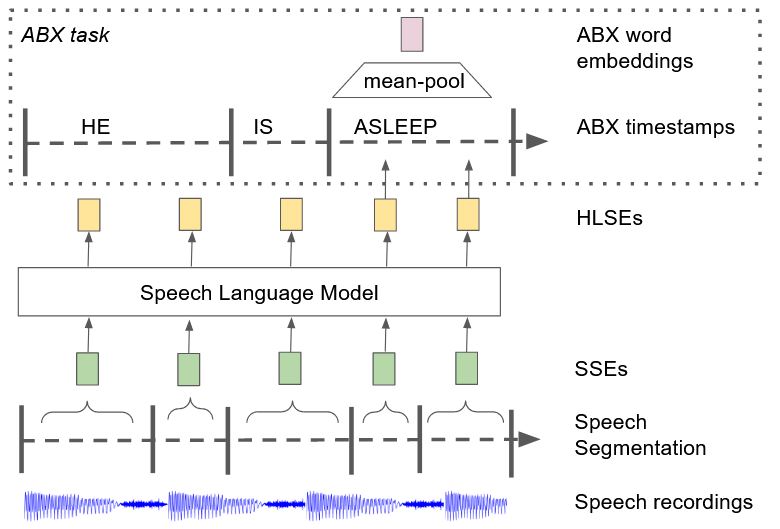}
    \end{center}
\caption{\textbf{Method for deriving HLSEs for semantic and POS evaluations}. Speech is segmented and converted into SSEs before being used to train a Speech LM. At test time, the HLSEs that overlap with the ABX timestamps by more than 40ms are mean-pooled to form the ABX word embeddings (here the word `asleep')}
     \label{fig:bert_schema}
\end{figure}

The idea here is to evaluate segmentation through its effect on a downstream language model. The evaluation is less direct than the segmentation metrics above, as it assumes that the segmentation is used to turn a spoken utterance into a series of fixed size SSEs, themselves used to train a continuous-input Speech Language Model. The metrics will therefore reflect each of these components (segmentation, SSE, Speech LM), but by keeping the SSE and Speech LM components constants, we can hope to study systematically the effect of the segmentation component. Our evaluation focuses on word-level representations, which we call here High-Level Speech Embeddings (HLSE). They are the speech equivalent of Word2Vec representations \cite{mikolov2012word2vec}, yet coming from an inexact segmentation. Here, we assume that a Speech LM has been trained on SSE inputs on a given dataset with a given segmentation and can be used at test time to provide embeddings associated to a set of test words. As our segmentation models work on whole utterances, we present the test words in continuous utterances and then mean pool the HLSEs from the Speech LM that overlap with that word to obtain a single vector. An overview of the building of generic HLSEs and their evaluation is shown in Figure \ref{fig:bert_schema}.

We introduce two zero-shot tasks to evaluate HLSEs that do not require training a further classifier, and are purely based on distances between these embeddings.  They are based on machine-ABX, a type of distance-based metric introduced in \citet{schatz2013abx} which computes a discrimination score over sets of $(A,B,X)$ triplets. For each triplet, $A$ and $X$ belong to the same category and $B$ plays the role of a distractor. The task for the model is to find the distractor in each triplet based on the cosine distance between the HLSEs of $A$, $B$ and $X$. The first task ($ABX_{sem}$), have $A$ and $X$ to be synonyms whereas $B$ is semantically unrelated to neither $A$ or $X$. The second task, $ABX_{POS}$, have $A$ and $X$ share the same POS-tags whereas $B$ has a different one. The triplets are all extracted from the Librispeech corpus training set. See appendix \ref{appendix:abx} for details on the construction of the triplets.

Another task, sSIMI from \citet{tuanh2021zerospeech}, also evaluates high-level representations of spoken words in a zero-shot and distance-based fashion. Yet, sSIMI differs from our ABX tasks in two important ways. First, sSIMI presents test words as pre-segmented chunk of speech without the context of the original sentence to help the Speech LM. Second, the task for the Speech LM is to predict a semantic similarity score given by human annotators to pairs of words. The Speech LM encodes the pre-segmented spoken words into HLSEs and the distance between the HLSEs is correlated with the human judgements. The correlation coefficient $r$ is used as the final sSIMI score.

\begin{figure}
     \begin{center}
     \includegraphics[width=\linewidth]{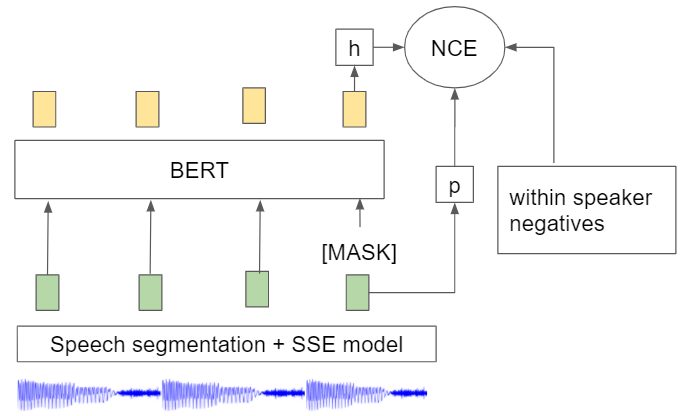}
    \end{center}
\caption{Our method to train a BERT model on SSEs. Speech is segmented into segments that are  converted into vectors by a frozen SSE model. BERT is trained with the NCE loss. $p$ and $h$ are used to project vectors into a common space to compute the NCE objective. The negative samples are random SSEs from the same speaker.}
     \label{fig:bert_training}
\end{figure}

\subsection{Datasets and Experimental settings}

The Zerospeech Challenge 2017 \cite{dunbar2017zerospeech} provides five corpora to evaluate speech segmentation systems. These corpora are composed of speech recordings from different languages split into sentences using Voice Activity Detection. Three corpora (Mandarin, 2h30; English, 45h; French, 24h) are used for development, and  two `surprise' corpora  for testing (German, 25h; Wolof, 10h). On each corpus, a separate SSE model from \citet{algayres2022sse} is trained and a separate run of DP-Parse is performed to produce a full-coverage segmentation. DP-Parse's hyper-parameters from Table \ref{table:hp}, as well as $q$'s formula (Equation \ref{eq:q}), were grid-searched to maximize token-F1 segmentation scores over the three development datasets. The two remaining test sets are used to show generalization of DP-Parse to new unseen languages. More details on hyper-parameters search in Appendix \ref{appendix:hp}.

For the ABX and sSIMI tasks, we train a BERT model as a Speech Language Model on the training set of the Librispeech dataset \cite{librispeech}, composed of 960 hours of English recordings. We proceed by first training a SSE model from \citet{algayres2022sse} on the Librispeech. Then, this latter dataset is segmented using DP-Parse. Each segmented speech sequence is aggregated into a single vector using the pre-trained SSE model. Segmented sentences are used to train a BERT model with masked language modelling and the Noise Contrastive Estimation (NCE) loss \cite{gutmann2021NCE} (see Figure \ref{fig:bert_training}). The BERT model is composed of 12 layers, 12 attention heads per layer with 768 neurons each and a FFN size of 3072 neurons. To compute the NCE, two heads $p$ and $h$ are composed respectively of one fully connected linear layer with 256 neurons and two fully connected layers with ReLU activation with also 256 neurons. Batches are composed of utterances from a single speaker, and the 200 negative samples for the NCE are chosen within the batch. 15\% of SSEs are masked in each batch during training. Only the BERT is trained, gradients are not propagated into the SSEs. As BERT is a multi-layer transformer model, the scores for ABX tasks and sSIMI are obtained by grid-searching the layer that performs best on the dev set of each task and evaluate that same layer on the corresponding test sets. The development and test sets for the ABX tasks are composed of triplets of sentences extracted from the training set of the Librispeech corpus. For sSIMI, the development and test set both contain two subsets: one with spoken words extracted from the Librispeech training set and one with spoken words synthesized with WaveNet \cite{oord2016wavenet}.

\begin{table}[t]
\centering
\resizebox{\linewidth}{!}{%
\begin{tabular}{ll}
\toprule
Minimum segment length  & $40ms$  \\ 
Maximum segment length  & $800ms$  \\ 
Nb. of segments in $L_0$ & $1$M\\ 
Concentration parameter $\alpha_0$ (eq. \ref{eq:dp}) & $100$\\ 
Nb. of neighbors in k-NN (eq. \ref{eq:density}) & $100$   \\ 
Lattice beam size & $10$  \\ 
Duration penalty $q$ (eq. \ref{eq:q}) & $\gamma=1.8$ \\  & $\delta=2$ (text)  \\
& $\delta=4$ (speech) \\
\bottomrule

\end{tabular}%
}
\caption{ DP-Parse hyper-parameters}
\label{table:hp}
\end{table}

\begin{table}[]
\centering\small
\begin{tabular}{l ccc}
\toprule
         &  Mandarin & French    & English \\
         &  (2h30) & (24h)    & (45h) \\\midrule 

DP Unigram  &    61m39s     &      240m44s & 62m11s    \\
DP-Parse  &    0m55s     &  25m12s & 62m11s    \\ \midrule
Speed-up factor  &    67.3     &      9.6 & 9.8    \\
\bottomrule
\end{tabular}
\caption{Time performances on 8 cpus for the standard configuration of DP-Parse (10 iterations) and the Dirichlet Process Unigram (2000 iterations) implemented by \citet{johnson2007ag}}
 \label{table:speed}
\end{table}

\section{Results}

\begin{table*}[]
\resizebox{\textwidth}{!}{%
\begin{tabular}{l rrrrrrrr rrrrrrr}
\toprule
 &\multicolumn{7}{c}{\bf Token F1 $\uparrow$}& &\multicolumn{7}{c}{\bf Boundary F1 $\uparrow$}\\ \cline{2-8} \cline{10-16}
&\multicolumn{3}{c}{\bf Dev} & & \multicolumn{2}{c}{\bf Test} & \multicolumn{1}{c}{}& &
\multicolumn{3}{c}{\bf Dev} & & \multicolumn{2}{c}{\bf Test} & \multicolumn{1}{c}{} \\ \cline{2-4} \cline{6-7} \cline{10-12}\cline{14-15}
\bf      Model              &  {Mand.}   & French &   {Engl.} & & German & {Wolof}  & {\textit{avg.}} & & {Mand.}   & French &   {Engl.} &  & German  & \multicolumn{1}{c}{Wolof}  & \multicolumn{1}{c}{\textit{avg.}}\\ 
\midrule[\heavyrulewidth] 
\multicolumn{13}{l}{\underline{\textit{Speech data}}}\\ 
Every 120ms        &    10.7 &   11.7 &    9.8   & &   5.4   &  12.4  &10.0   &  & 52.4	& 49.4  & 	48.2    &  &  41.3 & 55.0 &  49.2  \\
Syllables$^{\times}$ &     4.4 &    5.1	&    4.1	& &  3.0	  &   4.2  &{4.2}   &   & 49.1	& 46.3	&   43.7	&  &  38.1 & 40.6	& 43.6   \\
SEA$^+$              &    12.1 &    8.3	&    8.6 & 	&   7.5	  &  14.8  &{10.3}  &   & 52.2	& 48.4	&   46.1 &	&    40.9 & 54.2	& 48.4       \\
ES-Kmeans$^{\vee}$   &     8.1 &    6.30 &    19.2  & &\bf 14.5 &  10.9  &{11.8} &     & 54.5    & 43.3  & 	56.7  &  &\bf 52.3 &	52.8 &	51.9   \\
DP-Parse$^*$         &\bf 16.0 &\bf 15.3&\bf 21.9  & & 13.4	  &\bf 17.5 &{\bf 16.8}&  &\bf 59.3 & \textbf{55.9}	& \textbf{60.0} &	& 51.5	& \textbf{59.0}	& \textbf{57.1}     \\ 
\midrule
DP-Parse$^*$         & 28.2	& 30.9& 31.3 	& & 34.4 &	39.2 & {35} &	& 69.9	& 70.0	& 68.4	& & 64.4	& 75.5	& 69.4\\
$+$ weak sup. & \\
\midrule[\heavyrulewidth]
\multicolumn{6}{l}{\underline{\textit{Text data}}}\\ 
DP Unigram$^\ddagger$& 34.9	   &  57.0	& 64.7 &	& 33.6	& 38.8	&{45.7} &	& \bf{79.5}	& \bf{86.1}	& 88.9& &	71.6	& 77.3	& 80.7 \\
DP-Parse$^*$     & \bf 50.0 &	\bf  68.1 &	\bf  78.5 & &	\bf 67.4 &	\bf 69.1	& {\bf{66.6}} &	& 76.0 &	84.3 &	\bf 89.8 & &	\bf 83.5 &	\bf 84.1 &	\bf 83.5  \\
\midrule[\heavyrulewidth]
\end{tabular}
}
\caption{\textbf{Token and Boundary F1 scores} (the higher, the better) for speech and text segmentation models across  development and test set from the Zerospeech Challenge 2017 . The average scores are computed over the five datasets. *: ours, $\times$: \citet{okko2015sylseg}, +: \citet{bhati2020selfexpressing}, $\vee$: \citet{kamper2017eskmeans}, $\ddagger$: \citet{goldwater2009hdp}}
\label{table:f1}
\end{table*}

\subsection{Results on word-level segmentation}

Regarding text segmentation, DP-Parse compares favorably to the original Dirichlet Process Unigram model from \citet{goldwater2009hdp}. It produces a 21 points increase in token F1 compared to the original version (and 3 points increase in boundary F score) for a 10x runtime speed-up (Table \ref{table:speed}) under the Adaptor Grammar implementation of \citet{johnson2007ag}.\\

Regarding speech segmentation, we compare DP-Parse with the three best speech segmentation models submitted at the Zerospeech Challenge 2017 \cite{bhati2020selfexpressing,kamper2017eskmeans,okko2015sylseg}. Table \ref{table:f1} reports the token F1 and boundary F1 obtained by these models over the 5 datasets of the Zerospeech challenge 2017. Across all corpora, DP-Parse outperforms its competitors by at least 5 points in both boundary and token F1. We also introduce a naive baseline that draws boundaries every 120 milliseconds, disregarding the content of the speech signal. It turns out to be surprisingly competitive with all speech segmentation systems except DP-Parse. The latter is the only existing speech segmentation system that beats the naive baseline on all languages.

Most speech segmentation systems from the Zerospeech Challenge 2017 rely on off-the-shelf self-supervised representations of speech. The hope is that, without modification, these systems would benefit from future improvements in speech representations and mechanically lead to higher segmentation scores. Yet, such hopes have never been guaranteed by explicit experiments. Here, we test for the ability of DP-Parse to improve with better inputs. For that, we use the weakly-supervised version of the SSE model from \citet{algayres2022sse} trained with 10 hours of labelled speech data. On this type of input, DP-Parse doubles its token F1 score.


\begin{table*}[]
     \begin{center}

\resizebox{\textwidth}{!}{%
\begin{tabular}{llll cc c cc c cc}
\toprule
\bf Input                   & \bf  Segments   &\bf SSE  &\bf BERT     &\multicolumn{2}{c}{\bf Dev Set}&&\multicolumn{2}{c}{\bf Test Set} & &{\bf Dev Set} & {\bf Test Set} \\ \cline{5-6} \cline{8-9}\cline{11-12}
\bf Representations          & &  &  &\bf $ABX_{sem}$ $\uparrow$ &\bf $ABX_{POS}$ $\uparrow$ &&\bf $ABX_{sem}$ $\uparrow$ &\bf $ABX_{POS}$ $\uparrow$ & &sSIMI $\uparrow$ & sSIMI $\uparrow$\\\hline 
\multicolumn{4}{c}{\underline{\textit{No supervision, frame based}}}\\ 
CPC$^+$                   &-              & -      & -          &     51.22 [2] &	53.67 [2] &&	53.86 [2] &	54.68 [2] && \bf 6.14 [2]	& 4.34 [2]      \\
W2V2 base$^\times$        & -             & -            & -     &        54.96 [8]     &     55.78  [8]   &&   56.20 [8]      &        56.86 [8]   &&  3.85	[9] & 5.21 [9]      \\
HuBERT base$^\ddagger$    & -             & -            & -     &       54.65 [8]    &      55.65 [8] &&   55.12 [8]       &       56.77 [8] && 3.53 [8] &	0.68 [8]     \\
W2V2 large$^\times$       & -             & -         & -        &      57.69 [13]        &  \textbf{59.86 [13]}      &&   58.88 [13]        &       \textbf{60.93 [13]} && 2.04 [11] &	\bf 6.82 [11] \\
HuBERT large$^\ddagger$   & -             & -            & -     &      \textbf{58.04 [23]}        & 57.77 [23]     &&    \textbf{58.95 [23]}           &       57.19 [23] && 4.14 [7] &	0.55 [7]  \\ \hline 
\multicolumn{4}{c}{\underline{\textit{No supervision, segment based}}}\vspace{0.2em}\\ 
W2V2 base [8]$^\times$    & 120ms   & unsup$^\dagger$ & $\checkmark$ &     59.65 [7]       & \textbf{70.66 [7]}      &&   58.64 [7]          &       \textbf{71.68 [7]} && 4.10 [9]	& \bf 4.19 [9]  \\
W2V2 base [8]$^\times$    & Syllables$^\vee$   & unsup$^\dagger$ & $\checkmark$ &      59.09 [6]   &      68.58 [6]  &&  58.49   [6] &      68.81 [6] && 0.96 [9] &	3.48 [9]   \\
W2V2 base [8]$^\times$    & DP-Parse     & unsup$^\dagger$ & $\checkmark$ &         \textbf{61.09 [8]}     &      68.15 [7]    && \textbf{62.73 [8]}   
         &      69.35 [7] && \bf 4.54 [5] &	3.49 [5] \\\hline 
\multicolumn{4}{c}{\underline{\textit{Weak supervision, segment based}}}\vspace{0.2em}\\ 
W2V2 base [8]$^\times$    & 120ms    &  weak-sup$^\dagger$ & $\checkmark$ &    59.64 [8] &	70.04 [8]	&& 59.43  [8]	& 70.58 [8]     && \bf 5.83	[7] &  \bf 5.81	[7]  \\ 
W2V2 base [8]$^\times$    &Syllables$^\vee$ &  weak-sup$^\dagger$ & $\checkmark$ &   61.44 [6] &	70.89 [7]	 && 60.28 [6] &	71.03 [7]    && 0.51 [9] &	0.79 [9]  \\ 
W2V2 base [8]$^\times$    & DP-Parse      &  weak-sup$^\dagger$ & $\checkmark$ &  \bf 64.24 [8] &	\bf  72.65 [7] &&	\bf  64.82 [8] &	\bf  72.67 [7]  && 5.19 [9] &	4.51 [9]    \\\hline 
\multicolumn{4}{c}{\underline{\textit{Partial supervision, segment based}}}\vspace{0.2em}\\ 
W2V2 base [8]$^\times$    & gold words   &  unsup$^\dagger$ & $\checkmark$ &   67.89 [5] & 75.46 [7] && 	67.17 [5] &	75.61 [7]  &&  6.82 [5] &  	6.30 [5]      \\
W2V2 base [8]$^\times$    & gold words   &  weak-sup$^\dagger$ & $\checkmark$  & 71.38 [5] &	77.75 [6] &&	71.04 [5] &	77.64 [6] && 2.72 [5] &	 6.83 [5] \\ 
text                    & DP-Parse     & 1-hot      & $\checkmark$ & \bf 75.38 [4] &	\bf 78.57 [4]	&& \bf 74.89 [4] &	\bf 78.49 [4]	 && \bf 21.42 [2] & \bf 19.54 [2] \\
\hline
\multicolumn{4}{c}{\underline{\textit{Full supervision, segment based}}}\vspace{0.2em}\\ 
text                    & gold words    & 1-hot    & $\checkmark$   &       \textbf{83.23 [3]}         &  \textbf{81.07 [3]}      &&  \bf 84.12 [3]          &      \bf 80.09 [3] && \bf 41.78 [1] &	 \bf36.83 [1] \\
\bottomrule
\end{tabular}
}
\end{center}
\caption{\textbf{Semantic and Part-of-Speech discrimination scores} (the higher, the better). Segment-based models compute SSEs for each segment (obtained by speech segmentation) on which a BERT model is trained with a NCE loss. Frame-based models are pre-trained baseline models without speech segmentation nor BERT training. sSIMI scores are average over the Librispeech and synthetic subsets.
Layers used are given between brackets. $+$:\citet{oord2019cpc}, $\times$:\citet{baevski2020w2v2}, $\ddagger$:\citet{ning2021hubert}, $\vee$: \citet{okko2015sylseg}, $\dagger$: \citet{algayres2022sse}}
\label{table:abx_sem_pos}
\end{table*}
\subsection{Results on semantic and POS metrics}
In Table \ref{table:abx_sem_pos}, the \textit{segment-based} sections show how a BERT model can benefit from speech segmentation and SSE modelling on the tasks $ABX_{sem}$, $ABX_{POS}$ and sSIMI. To do that, we trained BERT models along the pipeline depicted in Figure \ref{fig:bert_training}.

Let us first analyse the scores on the ABX tasks. The section \textit{unsupervised, segment-based} shows that DP-Parse and two less-performant segmentation strategies lead to comparable ABX scores. Yet, the \textit{weak supervision} section shows that by improving the quality of the SSEs with weak supervision, DP-Parse performs higher than other segmentation methods. The \textit{partial supervision} section shows that with either perfect word boundaries or perfect segment representation (1-hot vectors), the ABX scores increase even more. Finally, the \textit{full supervision} is regular text-based LM that serves as a topline model.

As baseline systems, we propose \textit{frame-based} approaches that do not use speech segmentation nor SSEs. Speech-to-frames models like Wav2vec2.0, HuBERT or CPC \cite{baevski2020w2v2,ning2021hubert,oord2019cpc} are trained with masked language modelling in the spirit of text-based LM but on the raw speech signal. Therefore, these models should have already acquired some knowledge of semantics and POS-taggings. We evaluate speech-to-frames models directly on $ABX_{sem}$, $ABX_{POS}$ and sSIMI. To do that, we simply skip the Speech LM as well as the segmentation and SSE steps from the method in Figure \ref{fig:bert_schema}. Speech-to-frames models are used to encode the whole speech sentences from the ABX triplets, and the frames within the word boundaries provided by the ABX tasks are mean-pooled to form the HLSEs. The results show that Wav2vec2.0 Large and HuBERT Large are strong baseline systems for our ABX metrics. Yet, they score in average below the \textit{segment-based} approaches.

Regarding the scores obtained on sSIMI from \citet{tuanh2021zerospeech}, the table of results \ref{table:abx_sem_pos} shows important downsides on this task compared to our ABX tasks. First, the scores sometimes show large inconsistencies across development and test sets, which is not the case for the ABX tasks. Second, while our ABX tasks are sensitive to improvements in speech segmentation and SSE modelling, sSIMI is not and show comparable scores across most sections of Table \ref{table:abx_sem_pos}. One reason for that could be that self-supervised speech systems (Wav2vec2.0, HuBERT and CPC) as well as SSE models from \citet{algayres2022sse} are not trained to encode short sequence of speech, especially extracted as chunks from a sentence. 
  
\section{Conclusion and open questions}
We introduce a new speech segmentation pipeline that sets the state-of-the-art on the Zerospeech's datasets at $16.8$ token F1. The whole pipeline needs no specific tuning of hyper-parameters, making it ready to use on any new languages. We showed that the problem of speech segmentation can be reduced to the problem of learning discriminative speech representations. Indeed, using different level of supervision, our pipeline reaches up to $35$ token F1 score. Therefore, as long as the field of unsupervised representation learning makes headway, this method should automatically produce higher token F1 scores.

A first avenue of improvement of the current work is in the SSE component. Here, we took the system described in \citet{algayres2022sse} out of the box, and we showed good performance in speech segmentation compared to the state of the art, but there was still a large margin of improvement compared to text-based system. A recent unpublished paper \cite{kamper2022wordseg} came to our attention based on the non-lexical principle and showed similar or slightly better results than ours on a subset of the ZR17 language. \citet{kamper2022wordseg} also uses a segmentation lattice that resembles ours for inference. Yet, as we have shown, our system is monotonous with input embedding quality, and can therefore reach much better performance if the speech sequence embedding component were better trained. Further work is needed to improve the SSE component based on purely unsupervised methods.   

Despite our computational speedup, our current system would face challenges to scale up the approach to larger datasets than Librispeech. While it uses FAISS, an optimized search library, it is unclear that storing each instance of a possible segment is necessary. One interesting avenue of research would therefore be to downsample both instance lexicons to alleviate space complexity and/or to trade storage with compute by recalculating the embeddings on-line.

Another interesting improvement would be to compute bigram probabilities instead of unigram probabilities, as previously explored in \citet{goldwater2009hdp}. Yet, this possibility looks challenging as explained in appendix \ref{appendix:hdp}.

Finally, our method showed promising results in terms of semantic and POS encoding when taken as a preprocessor for a language model. Such phenomenon is displayed by our new in-context semantic tasks, but not by a previous without-context semantic task from \citet{tuanh2021zerospeech}. Further work is needed to show whether this can translate into high quality generations when used as a component to generative systems \cite{lakhotia2021gslm}.



\begin{table}[]
\centering
\resizebox{\linewidth}{!}{%
\begin{tabular}{lcc}\hline
 & \multicolumn{2}{c}{\bf Token-F1}  \\
\cline{2-3}
Languages & DP-Parse + $k$-means & DP-Parse + $k$-NN  \\ \hline
mandarin  & 0.1 & 0.147 \\
french    & 0.07 & 0.153 \\
english   & 0.08 & 0.219\\ \hline
\end{tabular}%
}
\caption{Token-F1 scores obtained by DP-Parse on three corpora from the Zerospeech 2017. The frequency estimation is performed by either a $k$-NN or by a $k$-means that knows the true number of clusters (found in the transcription of those datasets)}
\label{table:f1-kmeans}
\end{table}

\appendix

\section{Appendix}

\subsection{Construction of triplets for $ABX_{sem}$ and $ABX_{POS}$}
\label{appendix:abx}

Let's write the series of triplet from our ABX tasks: $(A_i,B_i,X_i)_{0\leq i<N}$. For all $i$, $A_i$ is defined as a tuple $(R_i^a,s_i^a,e_i^a,t_i^a)$ where $R_i^a$ is a recording of a whole sentence and $s_i^a$ and $e_i^a$ are the temporal boundaries of a word with phonetic transcription $t_i^a$. $B_i$ and $X_i$ are defined identically. The sentences are extracted from the Librispeech dataset, a 960 hours corpus of read English literature. 

The Speech LM is asked to encode the whole sentences $(R_i^a,R_i^b,R_i^x)$ and compute the three word embeddings for the words of interest using the provided timestamps. The ABX score is given by the following formula:
\begin{equation*}
    ABX(T,d)=\frac{1}{N}\sum_{i=0}^{N-1}\mathds{1}_{d(f_i^a,f_i^x) < d(f_i^b,f_i^x)}
\label{eq:abx}
\end{equation*}
where T is a collection of triplet word embeddings $(f_i^a,f_i^b,f_i^x)_{0\leq i<N}$ and $d$ is the cosine distance.\\
$ABX_{sem}$ is composed of 502 pairs (evenly split into a development and test set) of words created using a synonym dictionary. By sampling a list distractor for each pair, we reached 1557 different triplets. For each unique word found in the triplets, we sample 10 occurrences from the sentences of the Librispeech corpus. In total, $ABX_{sem}$ is computed over 1.5M triplets. $ABX_{POS}$ is composed of 9997 pairs of words built from the \textit{WordNet} database \cite{fellbaum1998wordnet}. The words in each pair have the same POS tags that can be either noun, verb or adjective. By sampling distractors and 10 occurrences of each word types, we reached 37.3M triplets.


\begin{table}[h!]
\resizebox{\linewidth}{!}{%
\begin{tabular}{l rrrr}
\toprule
& \multicolumn{4}{c}{\bf Token-F1}  \\
\cline{2-5}
         &  Mandarin & French & English & \textit{average} \\\hline 
$k=10$  &     16.8 &	14.7 &	21.5 &	\textit{17.7}  \\
$\bm{k=100}$  &   \bf  16.7 & 	\bf  15.3 &	\bf 21.9 &	\bf  \textit{18.0}   \\
$k=200$  &    16.9 &	15.4 &	21.9 &	\textit{18.0}   \\ \hline
$beam=5$  &   16.6	& 15.4	& 21.9	& \textit{18.0}   \\ 
$\bm{beam=10}$  &  \bf  16.7	& \bf 15.3	& \bf 21.9	& \bf \textit{18.0}   \\ 
$beam=100$  &    16.8	& 15.2	& 21.9		& \textit{18.0} \\ \hline
$\#L_0=300k$  &   16.9 &	14.7 &	21.5	&   \textit{17.7} \\ 
$\bm{\#L_0=1M}$  & \bf   16.7	& \bf  15.3	& \bf  21.9 & \bf \textit{18.0}   \\ 
$\#L_0=3M$  &   16.6 &	15.4 &	22.0 &	  \textit{18.0} \\ \hline
$\alpha_0=10^{-4}$  &   16.4 &	15.0 &	21.4 &	\textit{17.6}  \\ 
$\alpha_0=10$  &   16.3	& 15.3	& 21.7	& \textit{17.8}   \\ 
$\bm{\alpha_0=10^2}$  &   \bf 16.7	& \bf  15.3	& \bf  21.9	& \bf \textit{18.0}  \\ 
$\alpha_0=10^3$  &    17.0	& 15.1	& 21.7 &	\textit{17.9}  \\ 
$\alpha_0=10^5$  &  16.3 &	15.0 &	19.7 &	\textit{17.0}  \\ \hline
$\delta=1$  & 9.0 & 13.4 &	14.5 &	\textit{12.3}\\
$\delta=3$  &  17.0	& 14.8	& 21.7	& \textit{17.8} \\ 
$\bm{\delta=4}$  & \bf  16.7 & \bf 	15.3 &\bf 	21.9 &	\bf \textit{18.0}     \\ 
$\delta=5$  & 15.8 &	15.0 &	21.4 &	\textit{17.4}    \\
$\delta=9$  & 15.2 &	14.8 &	19.9 &	\textit{16.6} \\\hline
$\gamma=0$  & 14.7	& 14.7	& 19.9	 & \textit{16.4}  \\ 
$\gamma=1.6$  & 16.0 &	15.0 &	21.7	 & \textit{17.6}  \\ 
$\bm{\gamma=1.8}$  & \bf  16.7 &\bf 	15.3 &	\bf  21.9	 & \bf  \textit{18.0}   \\ 
$\gamma=2$  &  16.9 &	14.9 &	21.6 &	\textit{17.8}  \\
\bottomrule
\end{tabular}
}\caption{Token F1-score on the number of neighbors for the k-NN search ($k$), the concentration parameter ($alpha_0$), the beam-search size ($beam$), the number of samples in $L_0$ ($\#L_0$) and the pair ($\delta$,$\gamma$) from the penalty function. The default parameter values are $k=100$, $\alpha_0=100$, $beam=10$, $\#L_0=1M$, $\gamma=1.8$ and $\delta=4$ (also in bold in the table).}
 \label{table:gridsearch-hp}
\end{table}
\subsection{Ablation study} 

\subsubsection{k-means instead of $k$-NN}\label{appendix:kmeans}

To estimate DP-Parse parameters ($L_{0_w}$ and $L_w$), we followed a non-clustering approach with a $k$-NN density estimation. In this section, instead, we use k-means clustering to estimate DP-Parse parameters. We used the transcriptions of the development sets from the Zerospeech Challenge 2017 to give to the k-means the true number of clusters is it suppose to find when clustering SSEs from $L_0$ and $L$. The value of $L_{0_w}$ and $L_w$ are given by the size of the cluster in which $w$ is found. From Table \ref{table:f1-kmeans}, the segmentation scores obtained using k-means are much lower than those obtained by $k$-NN. Indeed, as shown in a study by \cite{algayres2020sse}, k-means is subject to the \textit{uniform effect} \cite{wu2021kmeans} which makes it not suited to estimate frequencies on highly skewed distributions, as the distribution of word types which follows the Zipf Law \cite{zipf}.

\subsubsection{DP-Parse hyper-parameters}\label{appendix:hp}

We provide in Table \ref{table:gridsearch-hp}, DP-Parse speech segmentation performances, as measured by token-F1 scores, over three datasets (Mandarin, English and French) for different hyper-parameters values. Surprisingly, increasing the number of 
neighbors ($k$), the number of samples in $L_0$ ($\#L_0$) or the beam size does not improve token-F1 scores. DP-Parse time complexity scales linearly with each of these three parameters, therefore we keep their values as low as possible.\\
Another unexpected result, is the low impact of the concentration parameter ($\alpha_0$). The value of this parameter should be a crucial as it controls the implicit vocabulary by controlling the amount of rare words in the segmentation. This observation has led us to create the penalty function $q$ from Equation \ref{eq:q} that offers a control on word-lengths, which also impacts the implicit vocabulary. This time, we noticed that $q$'s shape strongly impact token-F1 scores. By a careful tuning of the penalty function, we increased the token-F1 from $16,4$, with $\gamma=0$ (i.e. no penalty function) to $18$ ,with $\gamma=1.8$ and $\delta=4$. 

\subsubsection{Hierarchical Dirichlet Process}\label{appendix:hdp}

Goldwater's bigram Hierarchical Dirichlet Process (bigram-HDP), presented in \citet{goldwater2009hdp} for text segmentation computes the probability of two consecutive word-candidate, $P(\langle w_{i-1},w_i \rangle)$, to be segmented instead of only one, $P(w_i)$, as in the unigram dirichlet process (unigram-DP). In \citet{goldwater2009hdp}, the computation of the bigram-HDP probability $P(\langle w_{i-1},w_i \rangle)$ requires the number of different \textit{types} that can be found in all previously segmented bigrams with $w_i$ as second word. This is particularly difficult to adapt to speech segmentation with DP-Parse, because counting types requires an explicit clustering step. We doubt that using k-means would work as we have shown that clustering SSEs with k-means works poorly, even if the true number of clusters is known (see Table \ref{table:f1-kmeans}). More advanced clustering technics could work better than k-means as shown in \cite{kamper2014clustering}, e.g. Chinese Whispers Clustering, Hierarchical K-means or probabilistic clustering using GMMs, yet it would require a large effort to incorporate it to DP-Parse. \\


\bibliography{tacl2021}
\bibliographystyle{acl_natbib}

\clearpage\newpage

\setcounter{section}{0}
\setcounter{page}{1}
\setcounter{figure}{0}
\setcounter{table}{0}
\renewcommand{\thesubsection}{S\arabic{subsection}}
\def\thesubsectiondis{S\arabic{subsection}.}
\renewcommand{\thesection}{S\Roman{section}}
\renewcommand{\thefigure}{S\arabic{figure}}
\renewcommand{\thetable}{S\arabic{table}}
\renewcommand{\thepage}{\roman{page}}

\end{document}